\documentclass[sigconf]{acmart}

\usepackage{booktabs} 


\copyrightyear{2025}
\acmYear{2025}
\setcopyright{cc}
\setcctype{by}
\acmConference[ICVGIP 2025]{Indian Conference on Computer Vision, Graphics, and Image Processing}{December 17--20, 2025}{Mandi, India}
\acmBooktitle{Indian Conference on Computer Vision, Graphics, and Image Processing (ICVGIP 2025), December 17--20, 2025, Mandi, India}
\acmPrice{}
\acmDOI{10.1145/3774521.3774571}
\acmISBN{979-8-4007-1930-1/25/12}

\begin{document}
\title{PHAF: Personalized Hand Avatars in a Flash}
\titlenote{Produces the permission block, and
  copyright information}


  \author{Meghana Shankar}
\affiliation{%
	\institution{Samsung R\&D Institue Bangalore}
	\streetaddress{2870, Phoenix Building, Bagmane Constellation Business Park, Outer Ring Road, Doddanekundi Circle, Marathahalli}
	\city{Bangalore}
	\state{Karnataka}
	\country{India}
	\postcode{560037}
}

\author{Akanxit Upadhyay}
\affiliation{%
	\institution{Samsung R\&D Institue Bangalore}
	\streetaddress{2870, Phoenix Building, Bagmane Constellation Business Park, Outer Ring Road, Doddanekundi Circle, Marathahalli}
	\city{Bangalore}
	\state{Karnataka}
	\country{India}
	\postcode{560037}
}

\author{Anmol Namdev}
\affiliation{%
	\institution{Samsung R\&D Institue Bangalore}
	\streetaddress{2870, Phoenix Building, Bagmane Constellation Business Park, Outer Ring Road, Doddanekundi Circle, Marathahalli}
	\city{Bangalore}
	\state{Karnataka}
	\country{India}
	\postcode{560037}
}

\author{Green Rosh KS}
\affiliation{%
	\institution{Samsung R\&D Institue Bangalore}
	\streetaddress{2870, Phoenix Building, Bagmane Constellation Business Park, Outer Ring Road, Doddanekundi Circle, Marathahalli}
	\city{Bangalore}
	\state{Karnataka}
	\country{India}
	\postcode{560037}
}

\author{Pawan Prasad BH}
\affiliation{%
	\institution{Samsung R\&D Institue Bangalore}
	\streetaddress{2870, Phoenix Building, Bagmane Constellation Business Park, Outer Ring Road, Doddanekundi Circle, Marathahalli}
	\city{Bangalore}
	\state{Karnataka}
	\country{India}
	\postcode{560037}
}

\renewcommand{\shortauthors}{}

\graphicspath{{figures/}{pictures/}{images/}{./}}

\begin{teaserfigure}
	\includegraphics[width=\textwidth]{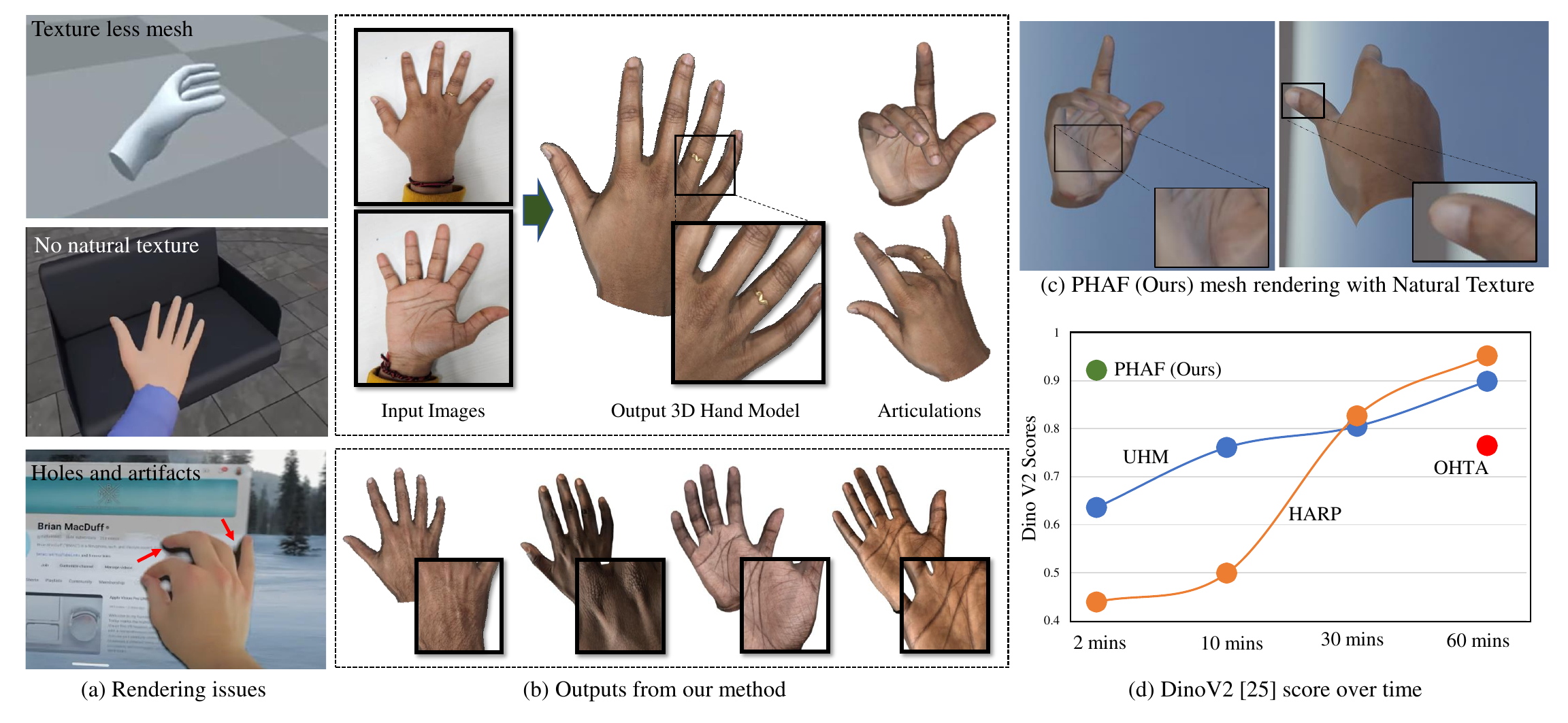}
	\caption{a) Rendering issues in existing HMDs - poor texture (row 1,2), holes (row 3 - red arrows). (b) Given 2 views of a target hand, our method (PHAF) synthesizes high visual fidelity renders under 2 minutes, preserving high frequency texture details. (c) High-quality renders, natural outputs from PHAF. (d) PHAF achieves comparable DinoV2 scores in significantly less time.}
	\Description{figure description}
	\label{fig:teaser_fig}
\end{teaserfigure}

\begin{abstract}
We present PHAF (Personalized Hand Avatars in a Flash ), a personalized photo-realistic hand avatar which provides high quality multi-view renders from just two images (dorsal and palmar views).Unlike slow optimization-based techniques, PHAF generates fast personalised textures for real-time deployment on edge devices. Our approach combines semantic guided mesh alignment and densified texture extraction to transfer high-frequency details efficiently. A view-based inpainting network refines textures ensuring smooth, continuous appearance. PHAF generalizes to novel viewpoints and leverages a parametric hand model for accurate articulations, making it compatible with standard graphics engines. Experiments show it is comparable to existing methods in visual fidelity while drastically reducing texture generation time by 30 $\times$, enabling practical AR/VR applications.

\end{abstract}

%
%


\maketitle

\section{Introduction}

Recent advancements in augmented and virtual reality (AR/VR) have led to a surge in VR head mounted devices (HMD) and applications, with a growing emphasis on personalization to enhance user engagement in virtual environments. Immersive technology, by definition, aims to fully immerse users in a digital world, making the experience as perceptually realistic as possible. A critical aspect of this immersion is the user’s ability to interact naturally with virtual objects, where the hands serve as the primary interface. Recent HMDs combine hand tracking with hand mesh rendering to enable the users to interact with the VR world using virtual hands. However, these methods often render simplistic hand meshes with minimal texture, impacting the realism of the interaction as shown in Fig. \ref{fig:teaser_fig}(a - row 1,2). Personalizing hand avatars to match the user's real world appearance has shown to significantly icrease engagement. The users feel a stronger sense of presence when they see their own hand features replicated in the virtual space \cite{xu2023harp}. There has also been methods which aims to provide a more natural experience to the users by overlaying a cut-out of the user's hand onto the VR environment. However, as shown in Fig. \ref{fig:teaser_fig}(a - row 3), such methods tend to generate artifacts due to erroneous segmentation of the user's hand, limiting realistic immersion. Further, this approach requires high resolution RGB cameras to be turned on during interaction, resulting in increased battery usage and over-heating. Hence, there is a need to generate articulatable and photo-realistic hand avatars with minimal RGB camera usage. This enables power saving along with seamless interaction experience for the users.  

Real-time generation of photo-realistic, animatable hand avatars remains an open challenge due to the complexity of hand geometry, articulation, and texture variation. Methods such as HandAvatar \cite{wang2023handavatar} and LISA \cite{corona2022lisalearningimplicitshape} proposed implicit modelling using Neural Radiance Fields (NeRF) \cite{mildenhall2020nerfrepresentingscenesneural} to encode hands as continuous neural functions. It enables photo-realistic novel-view synthesis without explicit mesh reconstruction. However these methods suffer from lack of control as pose and shape manipulation is non-trivial and each new pose requires expensive optimization, thus limiting practical applications. Recent methods such as Neural Volumes \cite{lombardi2019} improve dynamic scene modelling but still struggles with fine-grained deformations of the hand. They are also computationally expensive as rendering requires querying neural networks per pixel, preventing real time performance. To overcome the challenges of implicit representations, recent methods \cite{liu2025uhm,xu2023harp,zheng2024ohta,smith2024html,kim2024bitt} proposes geometry and texture refinement of explicit parametric hand models such as MANO \cite{manoref} and NIMBLE \cite{wei2022nimble}. For Example UHM \cite{liu2025uhm}and Harp \cite{xu2023harp} use phone scan videos to generate texture maps to create personalised avatars of the user. Though these methods generate high-fidelity custom textured models, they employ an iterative optimization approach using differential rendering, which adds significant time overheads and are slow for practical applications. Other methods such as HTML\cite{smith2024html} and BiTT \cite{kim2024bitt} use trained parametric appearance model for fast hand model generation. However, they often fail to generalize to unseen skin tones.

Despite significant advancements in hand modelling, we identify critical gaps between highly photo-realistic appearance generation and real-time applicability in AR/VR applications. 

We propose PHAF, an optimization-free method that generates high-quality textures in under 2 minutes, making it practical for AR/VR applications. Our method generates high quality, articulatable hand avatars from a pair of palmar and dorsal images of the hand (Fig. \ref{fig:teaser_fig} (b)). We propose a novel densified texture extraction module to directly transfer dense textures from the provided images to a parametric hand model. Our novel formulation eliminates the need for expensive iterative optimization, achieving 30 $\times$ improvement in run-time. We also propose a novel semantic guided geometric alignment module and a view based inpainting module to ensure that the generated texture map accurately fits the parametric models without artefacts. As shown in Fig. \ref{fig:teaser_fig} (b, c), our method generates high-fidelity hand avatars with detailed texture, outperforming the state-of-the-art methods. Further, it can be seen that our method generates hand avatar $\sim$ 30 $\times$ faster compared to state of the art with comparable quality (Fig. \ref{fig:teaser_fig}(d)). Our main contributions are summarized as follows:
\begin{itemize}
	\item We propose PHAF, a fast hand texture generation method with high frequency texture details, enabling practical deployment in AR/VR applications with only 2 views.
	\item We develop a texture synthesis approach by combining a semantically controlled geometric alignment with densified texture extraction to transfer hand appearance features to a parametric mesh.
	\item We further refine our texture using a view guided controlled inpainting network to render high quality novel and multi-view renders.
	\item Our framework is end-to-end developed for free-pose personalised photo-realistic hand avatars. We demonstrate that our approach achieves superior visual fidelity to existing methods with significantly reduced time for texture generation.
\end{itemize}

\section{Related Works}
\textbf{Parametric Foundations and Neural Hand Representations}. Parametric and musculoskeletal hand models \cite{SMPL_2015, scape,coregistration,pons} provide a structured foundation for realistic articulation. Nimble \cite{wei2022nimble} introduced a physically based representation with bones and muscles, enabling anatomically plausible motion with texture assets embedding. Extensions such as HTML \cite{smith2024html}  and Handy \cite{doe2024handy} combine these deformable geometries with texture mapping, offering consistent articulation with plausible appearance. However, such pipelines often depend on pre-scanned data or slow per-instance optimization to achieve personalization. 

Neural methods  \cite{jiang2022neumanneuralhumanradiance,peng2021animatableneuralradiancefields,peng2021neuralbodyimplicitneural} have augmented these geometric frameworks with learned volumetric appearance. Works like HandNeRF \cite{guo2023handnerf}  and HandOcc \cite{ivashechkin2025handocc}  produce striking photo-realism by encoding both geometry and view-dependent texture. Despite their fidelity, NeRF-style pipelines \cite{mihajlovic2022keypointnerfgeneralizingimagebasedvolumetric, park2021nerfiesdeformableneuralradiance, weng2022humannerffreeviewpointrenderingmoving}, remain computationally demanding and ill-suited for real-time AR/VR on edge devices. Hybrid strategies attempt to lower this barrier, HARP \cite{xu2023harp}captures detail from short RGB sequences, OHTA \cite{zheng2024ohta} and similar approaches achieves personalization from a single image. Yet, both methods still involve either iterative refinement or reliance on implicit priors, which can limit adaptability and speed. S2Hand \cite{chen2021modelbased3dhandreconstruction} and AMVUR \cite{ jiang2023probabilisticattentionmodelocclusionaware} presented a method to reconstruct both the appearance and geometry of a single hand from a single image. Nevertheless, their appearance is in blurred textures, omitting detail texture appearance.

Avatar systems such as XHand \cite{ gan2024xhand} address runtime constraints by prioritizing fast motion. While effective for expressive animation, they  require dense capture (multi-view or temporal) to maintain texture fidelity, making lightweight deployment challenging. On the other hand, DINAR \cite{ svitov2023dinardiffusioninpaintingneural} integrates neural textures with the explicit model for enhanced photo-realism but fall short in generating high-fidelity hand.

\noindent\textbf{High-Fidelity Texture Reconstruction and Generative Inpainting}. Parallel to these advances in geometry, a substantial body of work has focused on the appearance side reconstructing photo-realistic textures from limited inputs. HiFiHR \cite{ zhu2023hifih} demonstrates that high-frequency realism can be recovered from single RGB image but uses an iterative render-and-compare loop which makes it slow. DART \cite{ gao2022dart} expands personalization to include accessories, while UHM \cite{ liu2025uhm} streamlines capture to a single scan, though often at the cost of time. HandAvatar \cite{wang2023handavatar} enables free-pose rendering but can require lengthy capture sequences to achieve consistent textures. BITT addresses self-occlusion and interaction handling but at the cost of generalization.

To handle occluded or missing regions in sparse imagery, edge-aware projection and learning-based inpainting have proven valuable. EASI-Tex \cite{perla2024easi} refines projection boundaries for cleaner texture transfer, while Paint-it \cite{ lee2024paintit} and Paint3D \cite{ zeng2024paint3d} leverage generative and diffusion-based models for texture synthesis, offering stylistic flexibility but not necessarily domain-specific detail preservation. Broader texture-generation frameworks Mesh2Tex \cite{bokhovkin2023mesh2tex}, UniTEX \cite{liang2025unitex}, TEXTure \cite{richardson2023texture}, FlashTex \cite{deng2024flashtex}, Pix2Surf \cite{chen2023pix2surf}, as well as garment-oriented methods like Learning to Transfer Texture from Clothing Images to 3D Humans \cite{johnson2024texturetransfer}, FabricDiffusion \cite{wang2024fabricdiffusion}, and Garment3DGen \cite{patel2024garment3dgen}, highlight the potential of generative pipelines for complex surface texturing. Yet these methods largely target rigid or semi-rigid objects and seldom address the combination of skin realism, articulation, and real-time constraints required for hands.

\section{Proposed Methodology}

\subsection{Overview}
\begin{figure}[tb]
	\centering
	\includegraphics[width=\linewidth]{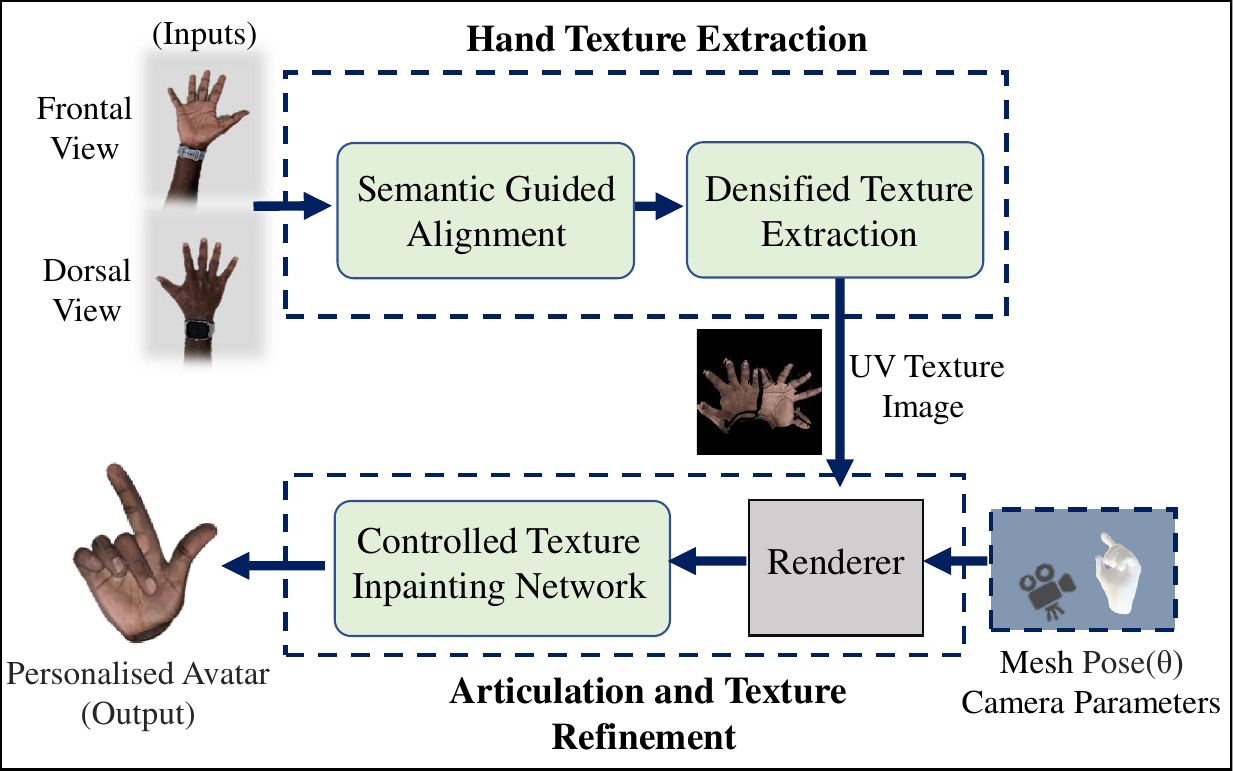}
	\caption{Our PHAF pipeline generates multi-view renders from just two images using novel semantic guided alignment module and densified texture extraction phase. The final missing textures are inpainted using a UNet based textured Inpainting network.}
	\label{fig: pipeline_fig}
\end{figure}

The PHAF pipeline, illustrated in Fig. \ref{fig: pipeline_fig}, begins with paired frontal and dorsal hand images as input and produces high visual fidelity renders for real-time articulations. At the core of our approach is a semantic-aware geometric alignment system that precisely maps image pixels to their corresponding mesh vertices through the underlying parametric model.  We also leverage the model’s geometry to obtain densified texture features and transfer it to a UV texture image. We address the inevitable missing texture regions through a specialized hand-specific Unet-based inpainting model \cite{hosen2022masked} trained exclusively on high-fidelity hand imagery. This neural inpainting stage fills missing regions and actively enhances texture details, recovering subtle high-frequency features that might be partially obscured in the original captures, while maintaining compatibility with standard animation pipelines.

\begin{figure*}[h]
	\centering
	\includegraphics[width=\linewidth]{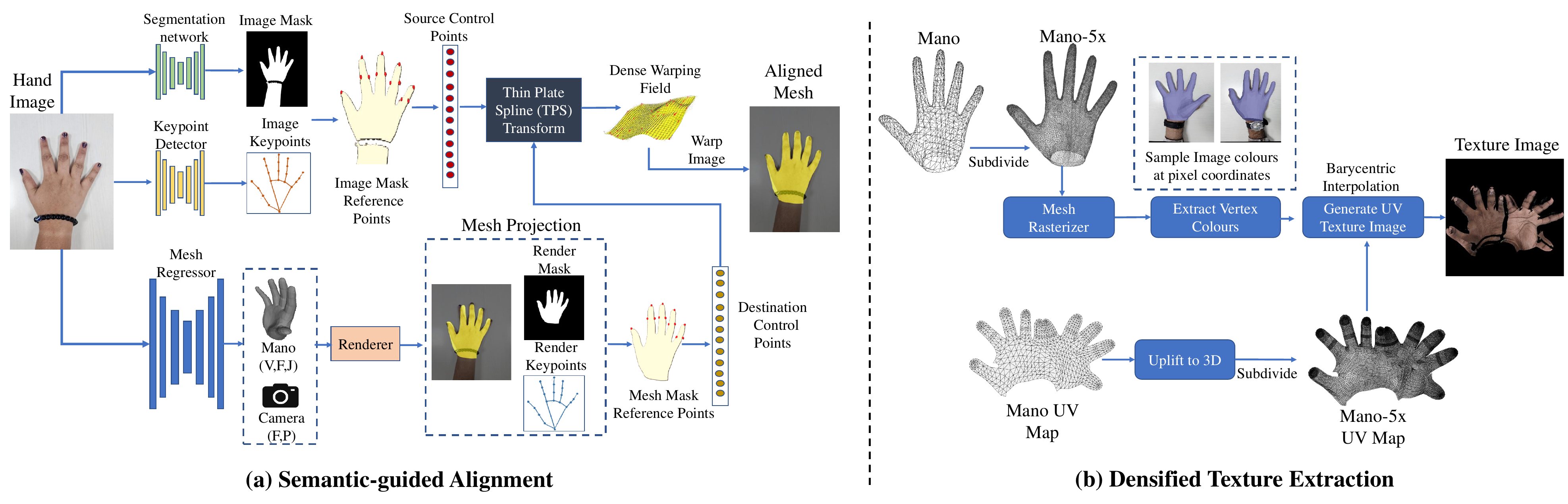}
	\caption{Our semantic guided alignment along with densified texture extraction extracts UV texture image from High density Mesh (Mano 5x). The texture images with high frequency visual fidelity is then used in real time rendering with a lower dimension mesh for fast rendering and the missing textures are inferred at runtime with our inpainting network.}
	\label{fig: segment_densification}
\end{figure*}

\subsection{Semantic guided alignment}
Precise texture mapping onto 3D hand meshes is essential for achieving realistic and personalized avatars. Traditionally, methods perform texture optimization via multi-view differential rendering, carefully matching color consistency across views. However, these approaches often assume that pixels sampled after rasterization inherently align semantically—i.e., correspond to the correct anatomical regions on the mesh. Hence they need multiple iterations to correct the colours. To overcome this, we introduce semantic-guided alignment prior to texture sampling. 

By identifying semantically relevant control points—such as fingertips, creases, and knuckle markers—on both the image segmentation mask and projected mesh silhouette, we establish accurate correspondences. These correspondences drive a Thin Plate Spline (TPS) \cite{tpsref} warping, ensuring the mesh is registered precisely to the image domain. With this warp, texture sampling captures anatomically correct pixels, preserving semantic fidelity in regions with critical identity markers like tattoos or moles.

Our novel semantic-guided geometric alignment module addresses 3 key aspects: (1) Unlike rigid alignment in \cite{xu2023harp} and \cite{liu2025uhm}, our method preserves local texture details through non-rigid deformation, hence overcoming iterative optimisation to get semantically correct textures, (2) Compared to neural warping approaches, we maintain explicit control over deformation through physically meaningful constraints and (3) Semantic guidance ensures alignment respects anatomical structures rather than just geometric features.

\begin{figure}[h]
	\centering
	\includegraphics[width=\linewidth]{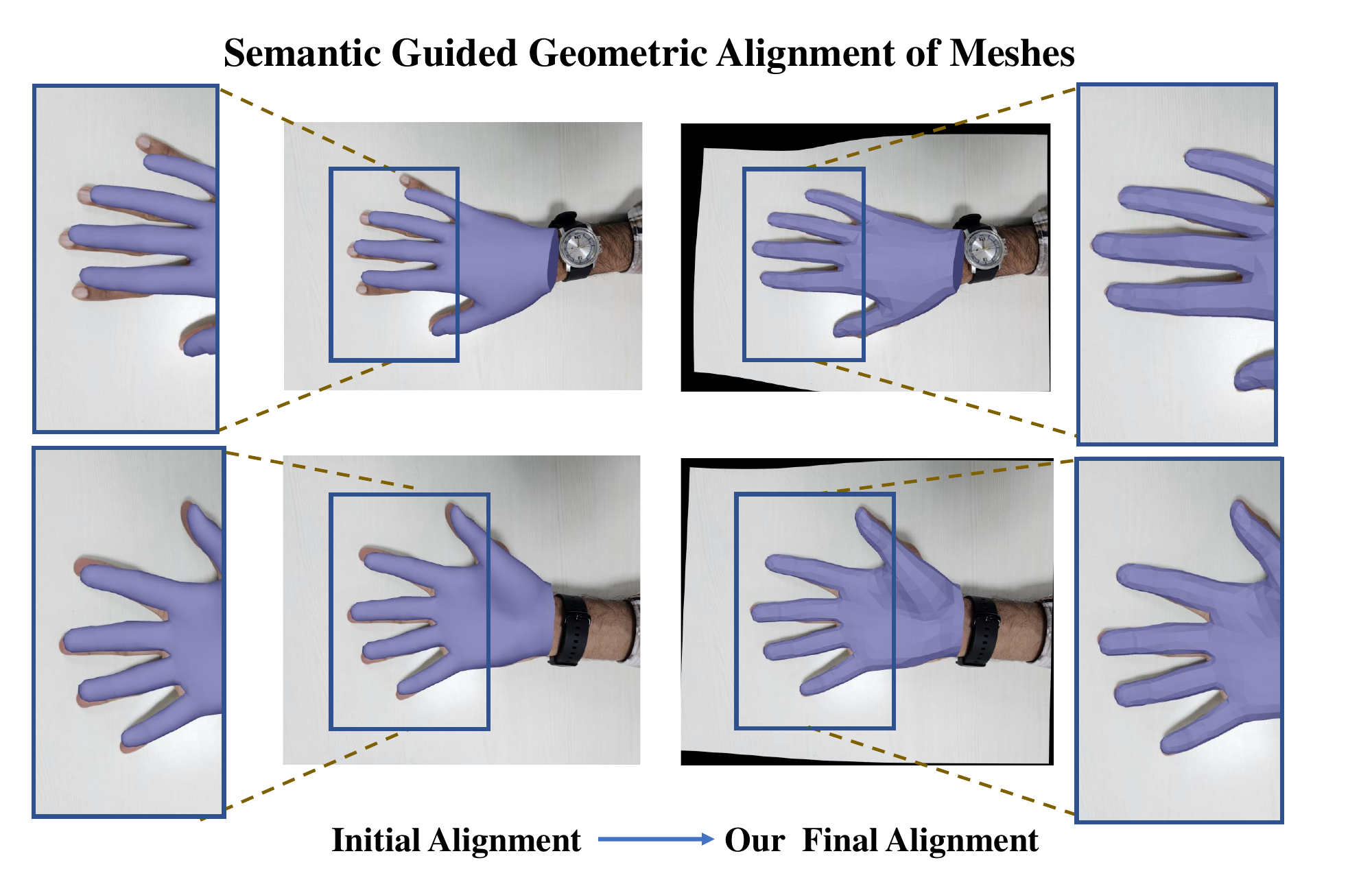}
	\caption{We showcase our semantic alignment feature, the incorrect mesh projection is corrected with our geometric alignment by warping the image to pass through the control points around the finger tips to have precise semantic matching for texture extraction.}
	\label{fig:alignment}
\end{figure}

Fig. \ref{fig: segment_densification} (a) illustrates our pipeline for Semantic guided alignment of the mesh with the hand image. For each image $I \in \mathbb{R}^{(HxWx3)}$, we extract two complementary representations: 21 hand keypoints $K^{img}$ using an existing hand keypoint detector and precise binary segmentation mask $S^{img}$ generated from a segmentation model. These semantic features capture both the structural topology and detailed contour information of the subject's hand.

Concurrently, we initialize a MANO \cite{manoref} parametric hand mesh $M(V, F, J)$, where $V$, $F$, and $J$ represents vertices, faces and joints respectively. It is then projected onto the image plane using estimated camera parameters. The projected mesh provides two crucial alignment references i.e. 2D keypoints $K^{mesh}$ obtained by projecting MANO's 3D skeleton, and a rendered silhouette mask $S^{mesh}$. We formulate the core alignment challenge as the positional discrepancy between image and mesh features.

\begin{equation}
	\epsilon_{align} = \frac{1}{n} \sum \left(K^{img} - K^{mesh} \right)^2
\end{equation}
where n = 21 keypoints. \\

Instead of aligning all the keypoints ($k_{p}^{img}$ , $k_{p}^{mesh}$),which results in higher time complexity, we choose a set of 14 control points for aligning $S^{img}$ and $S^{mesh}$ precisely at those points. These are the 5 finger tips, wrist keypoint and the DIP (distal interphalngeal) keypoints just below the tips.These points offer better alignment of semantic features than other points as confirmed with our experiments. As we want to align the silhouettes, these points are then extrapolated to the edges of the mesh contours as control points. As we want $S^{img}$ to align with $S^{mesh}$, we denote $X_p$ as the source control points and $Y_p$ as the destination control points lying on $S^{img}$ and $S^{mesh}$ respectively

We then establish anchored control point pairs $C_p={(X_p,Y_p)}$ where each control point $X_p$ corresponds to the image keypoint and $Y_p$ its mesh counterpart, and we employ a classic thin-plate spline (TPS) \cite{tpsref} model to define the warp $f:\mathbb{R}^2 \rightarrow \mathbb{R}^2$ . The TPS \cite{tpsref} model combines a global affine component with a sum of radial basis functions centred at the control points: 

\begin{equation}
	f(Z) = A * Z + t + \sum_{p=1}^{K} \left(w_p \phi \left( ||Z - C_p|| \right) \right) 
\end{equation}
where $\phi(r) = r^2 log r^2$. The unknown Affine parameters $A$,$t$ and kernel weights $w_p$ are solved by closed form linear equations. $Z$ is the query pixel coordinate. Hence $f(Z)$ gives the TPS \cite{tpsref} warper outputs, which is  the corresponding location in the warped image output for a query point $Z$. Hence this function smoothly interpolates the learned nonrigid alignment from the defined control correspondences as seen from the Fig. \ref{fig:alignment}.

\begin{figure}[h]
	\centering
	\includegraphics[width=0.9\linewidth]{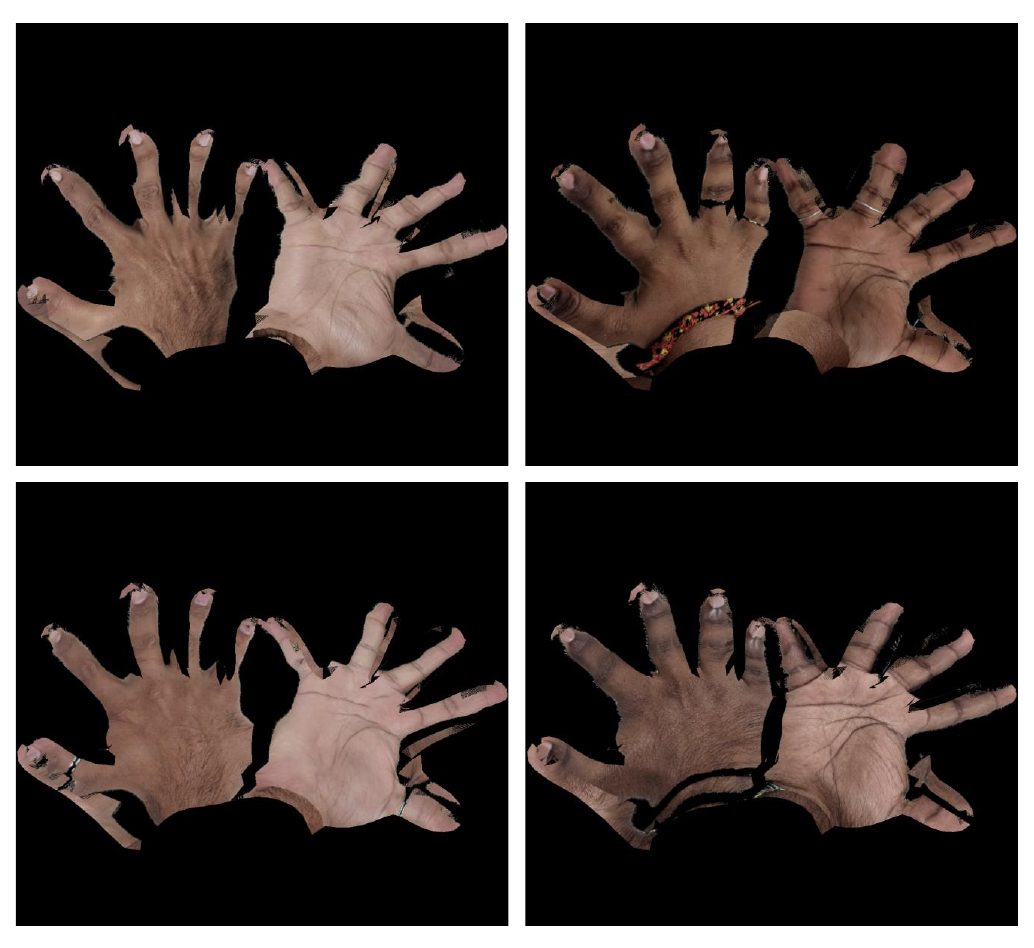}
	\caption{Extracted UV Texture images of the subjects: The unwrapped texture images consists of missing textures around the hand sides and in-between fingers due to occluded views.}
	\label{fig: texture_maps}
\end{figure}

\subsection{Densified Texture Extraction for High-Fidelity Hand Avatars}
Low-dimensional meshes with limited vertices cannot capture sufficient color detail when projected onto images, because each vertex samples only one color and large triangles lead to coarse, blurred textures. As its known that vertex colors get only one color per vertex and color is then interpolated inside the triangles, it causes loss of fine texture detail where the geometry is sparse. To overcome this, we densify the mesh by subdividing the base topology, thereby increasing vertex count and enabling more granular sampling of image pixels during rasterization. More vertices translate directly to more sampled colors and improved high-frequency content capture during texture projection. Hence densification of mesh allows for richer, perceptually accurate texture extraction critical for personalized hand avatars.

We first take the low-resolution MANO \cite{manoref} mesh and apply a $5\times$ subdivision scheme to obtain a very dense $\sim$ 700K vertices structure. The subdivision process follows:
\begin{equation}
	V_{dense} = SS(V_{MANO})
\end{equation}

where $SS$ is the subdivision operator applied iteratively to refine the mesh topology, $V_{MANO}$ and $V_{dense}$ represents the vertices of MANO and subdivided MANO respectively. With the densified mesh $V_{dense}$ and the warped input image $I$, we perform per-vertex color sampling using fragment shader-based rendering. To extract vertex colors, each vertex $v_i \in V_{dense}$ is projected onto $I$ via the previously estimated camera matrix $P = K \left[R | t\right]$.

\begin{equation}
	\left[ {U_i \atop V_i} \right] = \pi (P v_i) 
\end{equation}

where $\pi$ denotes perspective division, $(U_i, V_i)$ represents the image coordinates onto which $v_i$ is projected.

A fragment shader processes each triangle, interpolating UV coordinates and fetching the corresponding RGB value from Image $I$. For each triangular face $f=(V_a,V_b,V_c)$, we compute the interpolated color $C_f$ using barycentric weights $(\alpha, \beta, \gamma)$:

\begin{equation}
	C_f = \alpha C(V_a) + \beta C(V_b) + \gamma C(V_c)
\end{equation}

This ensures consistent colouring across the dense mesh, with natural transitions aligned to the image's photo-metric structure. However, this high-vertex-count mesh—while accurate—poses challenges for real-time rendering. To maintain performance, we transfer the dense coloration to a texture image that can be sampled by a low-resolution MANO \cite{manoref} mesh. While we do not have direct UV coordinates for the subdivision vertices, we utilize the fact that subdivision preserves topological mapping. By treating the UV layout of the low-resolution mesh as a 2D plane embedded in 3D where the MANO UV coordinates $UV_{MANO} \in \mathbb{R}^2$ are lifted to 3D space by appending a zero z-coordinate.

where $UV_{3D} = (UV_u, UV_v, 0)$. 

We then apply the same categorical subdivision to this UV plane. This creates a refined UV grid that mirrors the high-resolution mesh's vertex topology but resides in UV space,ensuring a bijective mapping between high-res 3D vertices and UV coordinates. Using UVdense, we unwrap the vertex colors onto a 1024x1024 texture image. The result is a continuous, high definition texture image $T(u,v)$ representing the dense surface coloration baked from the aligned RGB image. Fig. \ref{fig: texture_maps} shows some of our extracted UV texture images. This texture image can then be  sampled by a low resolution mesh which allows for high frequency texture renders. 

\subsection{Controlled Inpainting for Side and Inter-Finger Texture Completion}
\label{subsection: inpaint_network}
We address texture gaps—particularly on mesh sides and between fingers by applying controlled inpainting using a Residual Attention U-Net, inspired by \cite{hosen2022masked}. This architecture introduces residual blocks for stabilizing training and mitigating vanishing gradients, alongside attention units to focus computation on structurally critical masked regions.
Instead of naive interpolation, our model is trained on real data simulating missing textures. We create training inputs by performing morphological erosion at hand mask contours and overlaying random elliptical black blotches, mimicking occluded or unprojected areas.

The network is optimized with a composite loss:
\begin{equation}
	L_{total} = \lambda_{1} L_{L1} + \lambda_{2} L_{L2} + \lambda_{3} L_{COS} + \lambda_{4} L_{HSV}
	\label{equation:loss_fn}
\end{equation}

where $L_{L1}$ and $L_{L2}$ penalizes absolute and squared color differences respectively. $L_{COS}$ enforces perceptual feature alignment via cosine similarity, and $L_{HSV}$ ensures hue consistency in the color space. The residual blocks mitigate vanishing gradients, while attention mechanisms reduce blurring, critical for high-resolution hand textures. Our experiments confirm that the HSV loss significantly reduces color drift, and the elliptical mask augmentation improves robustness to real-world occlusions.

This controlled inpainting pipeline yields semantically coherent, high-fidelity texture completion, ensuring anatomically consistent transitions across occluded regions. The attention mechanism precisely targets challenging areas, while residual pathways uphold structural consistency, delivering realistic side and inter-finger textures for personalized hand avatars.

\section{Experiments}
\subsection{Experimental Setup and Data Sources}
To evaluate our pipeline, we begin by extracting initial MANO \cite{manoref} mesh parameters and camera poses using the InterWild \cite{moon2023interwild} hand modeling applied to input images. This yields pose, shape, and camera intrinsics and extrinsics suitable for projecting our mesh reliably in alignment with the image domain. Concurrently, we detect 21 hand keypoints using MediaPipe Hands and compute segmentation masks via the Segment Anything Model (SAM) \cite{kirillov2023segany} on the same input frames. These two semantic inputs form the foundation for our contour aware TPS \cite{tpsref} alignment stage.Using the recovered mesh parameters and camera pose, we rasterize the MANO \cite{manoref} mesh into image space to generate a mesh silhouette mask $M_{mesh}$ . This mask provides ground truth geometry projection which we use in combination with SAM \cite{kirillov2023segany} masks in the alignment step, ensuring the warped image mask matches the rendered mesh silhouette to sub pixel precision.

\noindent\textbf{Lab-Captured Multi-View Hand Scans}:
We captured our own sample data of 15 subjects, 5 poses each, 50K images, with High-resolution (12MP) RGB phone captures with diversity in skin tones, gender and accessories. We use this to train our inpainting network and to evaluate the visual fidelity of the photo-realistic avatars.

\noindent\textbf{HARP \cite{xu2023harp} Dataset}:
We used all 5 sequences shared by the researches which include Subject\_1, smile, smile\_black, dimlight and tattoo .We use this data to evaluate the quantitative and qualitative performance of the photorealistic avatars.

\noindent\textbf{Hands11K \cite{afifi201911kHands} Dataset}:
11k Hands dataset, a collection of 11,076 hand images (1600 x 1200 pixels) of 190 subjects, of varying ages between 18 - 75 years old. Each hand was photographed from both dorsal and palmar sides with a uniform white background  with metadata on subject ID, gender age , skin color etc but we discard the metadata and use only the images for our training.

\subsection{Training the Inpainting network}
\begin{figure}[h]
	\centering
	\includegraphics[width=\linewidth]{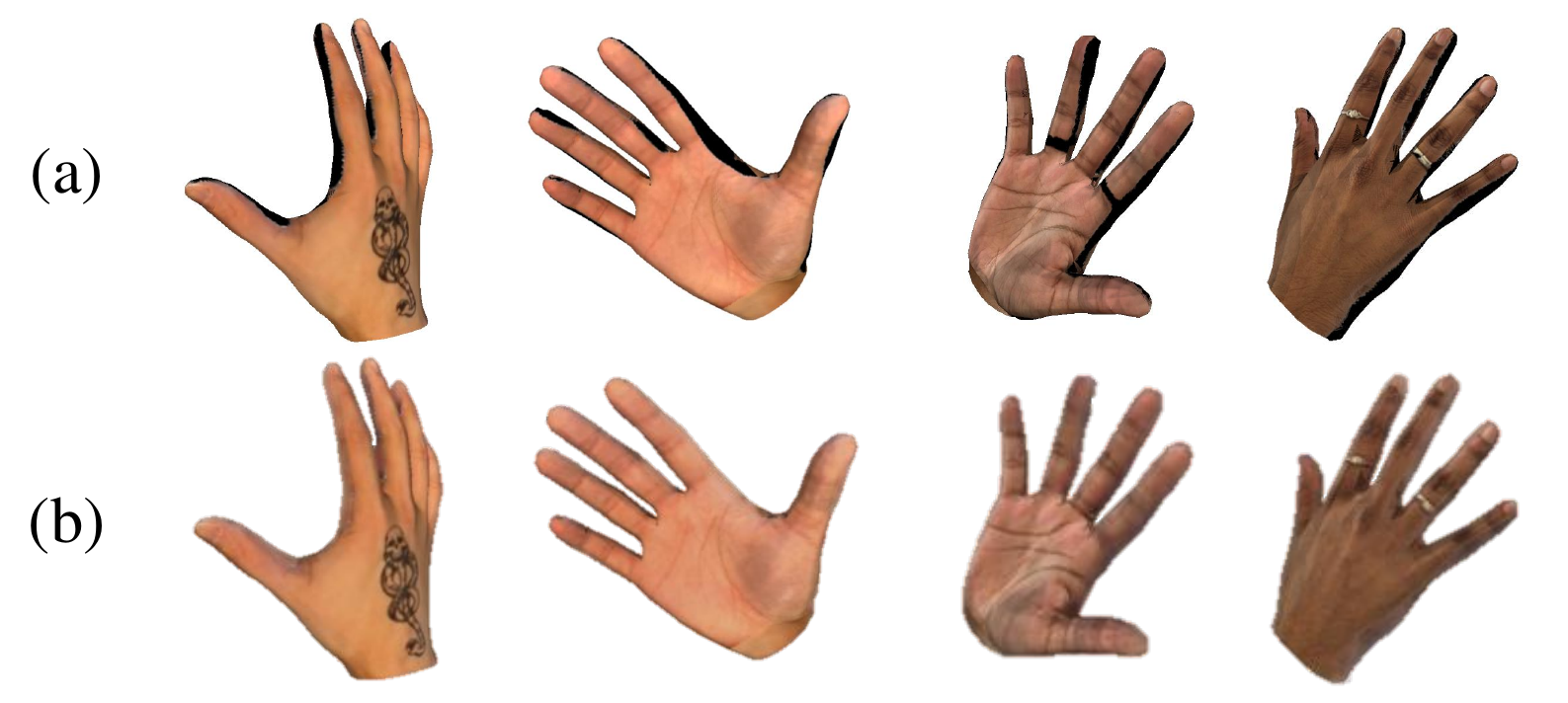}
	\caption{Qualitative results for the evaluation of the trained Inpainting network. Row (a) is the input to the network which is having several artifacts. Row (b) shows the inpainted results from the network.}
	\label{fig: qual_network}
\end{figure}

In order to enable robust hole filling and artifacts removal, we trained the U-Net based inpainting network described in Section \ref{subsection: inpaint_network}. We combined Hands11k \cite{afifi201911kHands}, HARP \cite{xu2023harp} and our Lab captured hand scans to form a dataset of $\sim$ 45K hand images along with their corruption masks and trained the network in a self-supervised manner. The dataset was randomly divided into training and testing sets of $\sim$ 40K and $\sim$ 5K images respectively. Based on varied sets of experiments, we identified suitable weightage for each term in the loss function defined in Equation \ref{equation:loss_fn}, and those were $\lambda_{1} = 1.0$, $\lambda_{2} = 0.5$, $\lambda_{3} = 0.3$ and $\lambda_{4} = 0.3$. The model was trained for 80 Epochs on an Nvidia RTX A6000 GPU. We also evaluated the model on the testing set consisting of $\sim$ 5k images both Quantitatively as well as Qualitatively. In terms of quantitative evaluation, we got a PSNR of $28.32$, SSIM of $0.968$ and an LPIPS of $0.0623$. Fig. \ref{fig: qual_network} showcases the results of our qualitative evaluations.

\subsection{Evaluation Metrics}
To evaluate realism and personalization of generated hand textures we adopt ViTScore \cite{vitref} and DINOv2 \cite{dinov2ref} cosine similarity as our principal metrics. These are built on Vision Transformer \cite{vitref} embeddings and designed to capture image semantic similarity and outperforms classical pixel/structure-based measures such as PSNR, SSIM etc for semantic tasks. While LPIPS, though perceptually motivated, lacks the specificity to differentiate unique identity markers.

\vspace{-0.4cm}
\begin{figure}[h]
	\centering
	\includegraphics[width=\linewidth]{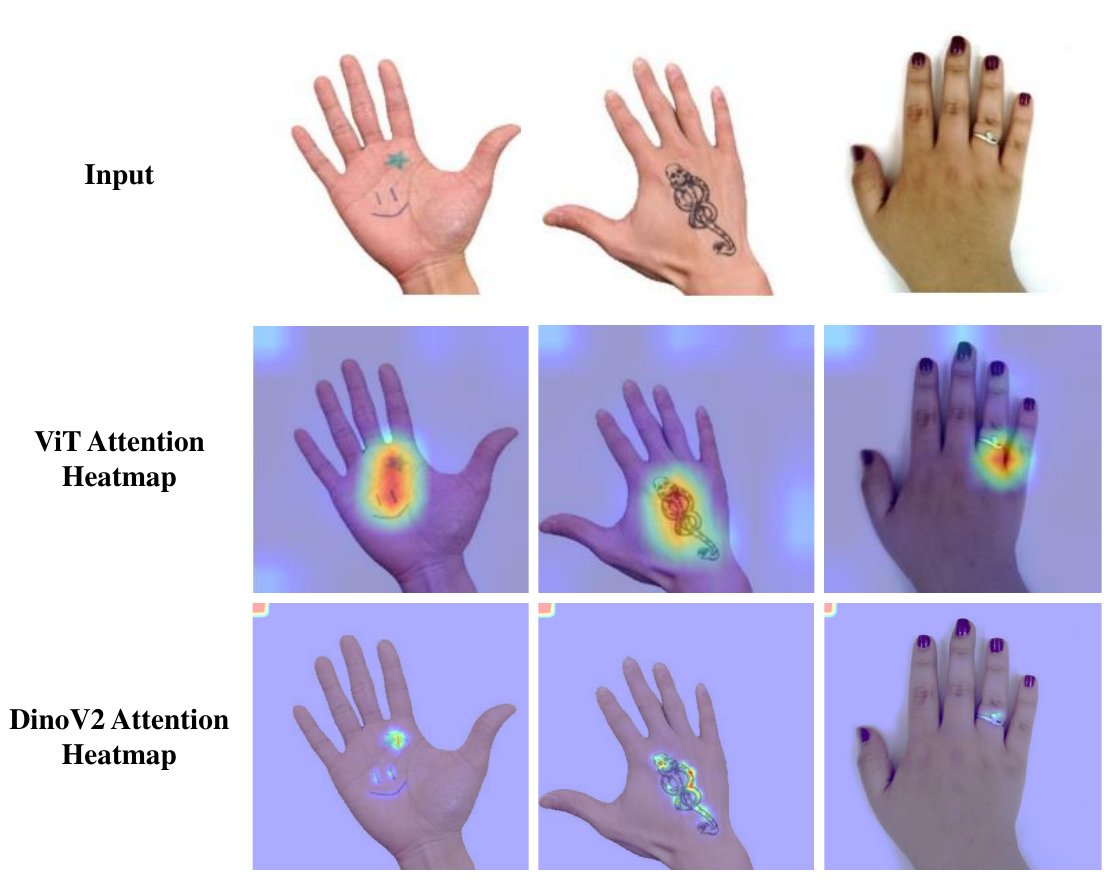}
	\caption{DinoV2 \cite{dinov2ref} and ViT \cite{vitref} heatmaps shows that these metrics attend to user specific identifiers which are crucial for personalised hand avatar}
	\label{fig: heatmaps}
\end{figure}

\begin{table*}[!htbp]
	\caption{Quantitative results against state-of-the-art methods using ViT \cite{vitref} and DINOv2 \cite{dinov2ref} scores. Our method significantly outperforms other methods at similar runtime}
	\resizebox{\textwidth}{!}{%
		\huge
		\begin{tabular}{@{}c|cc|cc|cc|cc@{}}
			\toprule
			\textbf{\begin{tabular}[c]{@{}c@{}}Time   since\\      optimization started\end{tabular}} & \multicolumn{2}{c|}{\textbf{2   mins}} & \multicolumn{2}{c|}{\textbf{10   mins}} & \multicolumn{2}{c|}{\textbf{30   mins}} & \multicolumn{2}{c}{\textbf{60   mins}} \\ \midrule
			\textbf{Methods} & ViT Score $\uparrow$ & DINOv2 Score $\uparrow$ & ViT Score $\uparrow$ & DINOv2 Score $\uparrow$ & ViT Score $\uparrow$ & DINOv2 Score $\uparrow$ & ViT Score $\uparrow$ & DINOv2 Score $\uparrow$ \\ \midrule
			UHM \cite{liu2025uhm} & 0.712 & 0.636 & \textbf{0.859} & \textbf{0.761} & \textbf{0.892} & 0.805 & \textbf{0.932} & 0.899 \\
			HARP \cite{xu2023harp} & 0.665 & 0.439 & 0.688 & 0.499 & 0.799 & \textbf{0.827} & 0.896 & \textbf{0.952} \\
			OHTA \cite{zheng2024ohta} & - & - & - & - & - & - & 0.491 & 0.765 \\
			\textbf{PHAF (Ours)} & \textbf{0.864} & \textbf{0.922} & - & - & - & - & - & - \\ \bottomrule
		\end{tabular}
	}
	\label{table: quant_res}
\end{table*}

\begin{figure*}[!htbp]
	\centering
	\includegraphics[width=\linewidth]{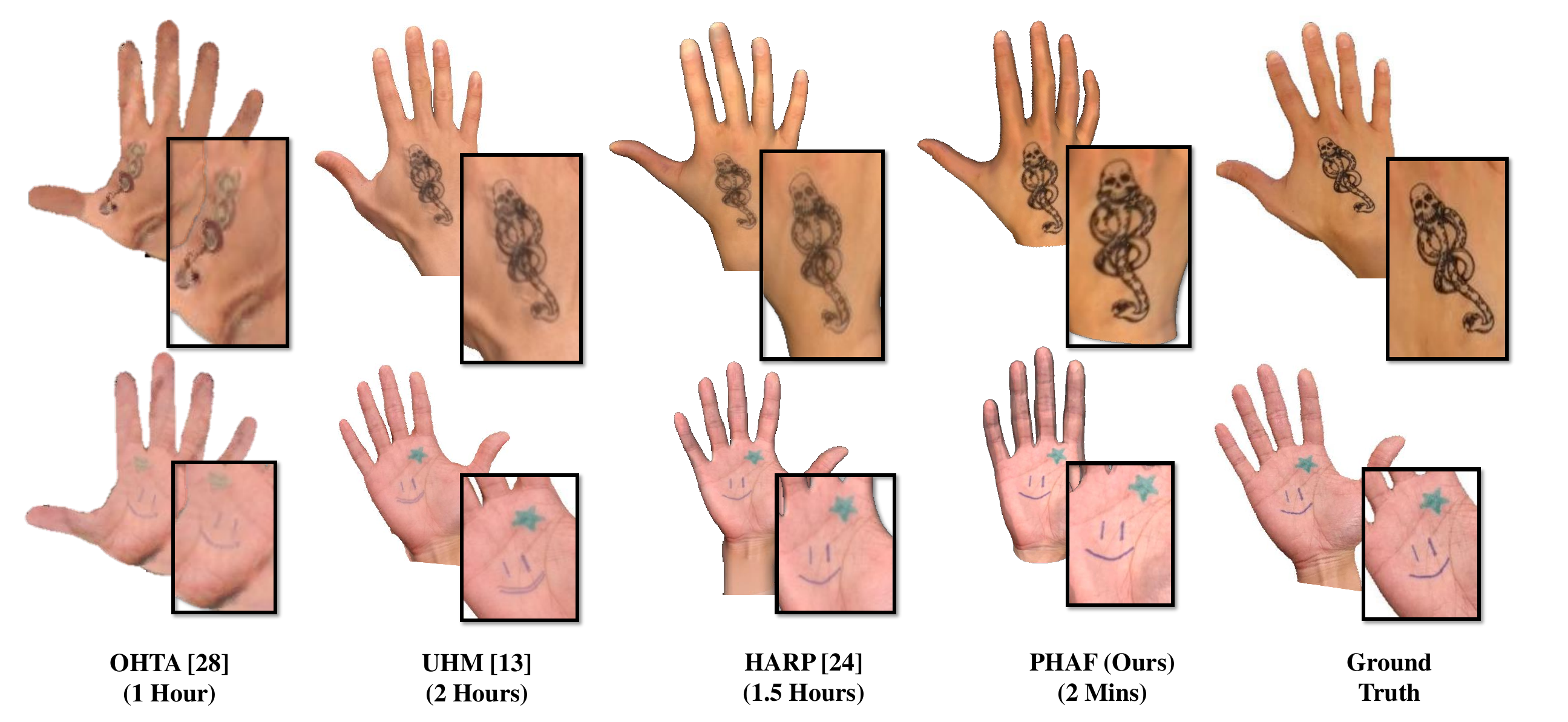}
	\caption{Qualitative comparisons of the proposed PHAF with UHM \cite{liu2025uhm}, HARP \cite{xu2023harp} and OHTA \cite{zheng2024ohta}. PHAF achieves }
	\label{fig: qual_results}
\end{figure*}

Beyond raw scores, a key advantage of transformer-based embeddings is interpretability: attention heatmaps from ViT/DINO models consistently localize discriminative patches and can be visualized to verify that the metric attends to user-specific cues (e.g., tattoos, moles, creases)—i.e., personalized identifiers that conventional metrics overlook. Fig. \ref{fig: heatmaps} presents attention heatmaps from both DINOv2 \cite{dinov2ref} and ViT \cite{vitref} generated at intermediate self-attention heads. This targeted localization underscores the strength of transformer-based embeddings in capturing fine-grained, identity-relevant texture details—capabilities that traditional metrics like SSIM, PSNR, or LPIPS cannot offer.

\subsection{Quantitative Evaluation}
We evaluate our method, PHAF, against three state-of-the-art personalized hand texture techniques using the HARP \cite{xu2023harp} dataset for benchmarking, this data features video scans of five subjects—four with distinct skin markers (e.g., tattoos) and one without any markers. HARP \cite{xu2023harp} reconstructs personalized hand avatars from short monocular RGB videos using an explicit mesh-based model with vertex displacement, normal, and albedo maps, coupled with a shadow-aware differentiable renderer. UHM \cite{liu2025uhm} adapts a Universal Hand Model via a quick phone scan, producing high-fidelity, animatable avatars through a combined tracking-and-modeling pipeline . OHTA \cite{zheng2024ohta} — a single-image neural rendering approach that attempts one-shot avatar creation but struggles to capture fine person-specific texture details.

Using our transformer-based metrics (ViTScore \cite{vitref} and DINOv2 \cite{dinov2ref} cosine similarity) as shown in Table \ref{table: quant_res}, PHAF achieves comparable semantic texture similarity to HARP \cite{xu2023harp} and UHM \cite{liu2025uhm}, even though our pipeline runs in just $\sim$ 2 minutes, versus $\sim$ 1 hour for the video-based methods, representing a 30 $\times$ speed-up. Also, PHAF significantly outperforms all the methods at similar run-time (2 mins). OHTA \cite{zheng2024ohta} fails to achieve acceptable results even after extended processing ($\geq$ 1 hour), particularly for subjects with prominent skin markers. These findings underscore PHAF’s balance of efficiency, accuracy, and personalization, outperforming one-shot implicit methods and matching video-optimization techniques—with a 30x reduced time cost.

\begin{figure}[h]
	\centering
	\includegraphics[width=\linewidth]{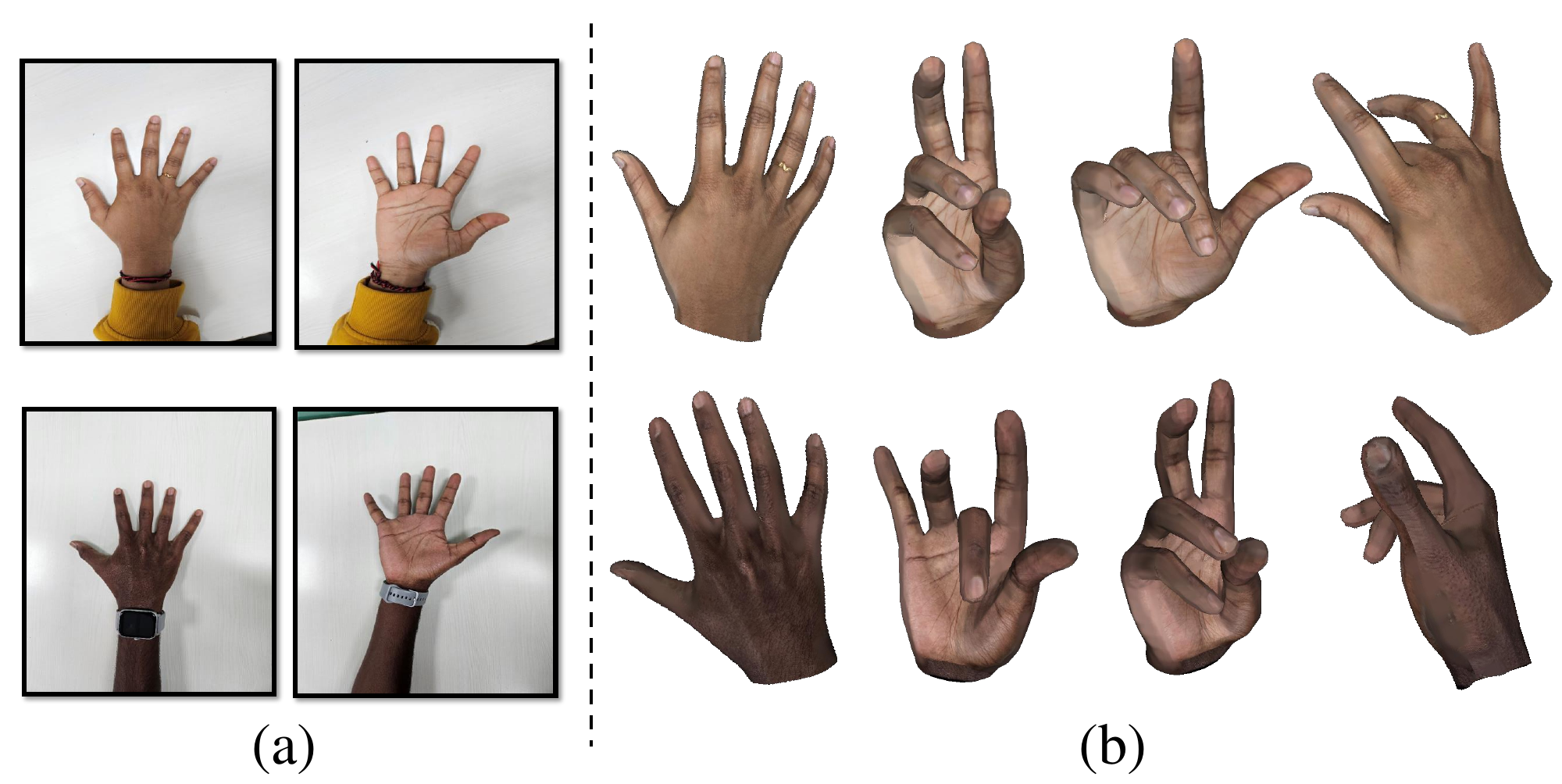}
	\caption{(a) shows the input Palmar and Dorsal images. (b) shows various articulation of the generated textured meshes using PHAF.}
	\label{fig: articulation}
	
\end{figure}
\subsection{Qualitative Evaluation}
We qualitatively compare our hand-textured avatar-PHAF against leading personalized hand avatar methods—UHM (Universal Hand Model) and HARP, both using video input, as well as OHTA, a state-of-the-art neural rendering approach from a single image. UHM achieves high fidelity via short phone scans and explicit mesh adaptations; HARP reconstructs personalized geometry and appearance from a short monocular video through optimizable mesh-based models and differentiable rendering; OHTA generates animatable hand avatars from one image using data-driven implicit priors.

In Fig. \ref{fig: qual_results}, our avatar demonstrates perceptually superior texture fidelity: fine details such as tattoos and unique skin markers appear distinctly and consistently—details that are often blurred or absent in the comparative methods. The sampling from a $\sim$ 700k vertex dense mesh helps to preserve details beyond what baseline methods achieve using normal maps and vertex displacements. Notably, our method completes the entire pipeline in approximately 2 minutes, in stark contrast to the $\sim$ 1-hour runtime of the video-based methods, offering a 30$\times$ acceleration without compromising visual realism.

Fig. \ref{fig: articulation} illustrates articulated mesh renderings. Since our pipeline leverages the standard MANO \cite{manoref} mesh, it supports diverse hand articulations and integrates seamlessly with existing research infrastructures—eliminating the need for custom geometry as required by other approaches.

\section{Ablation Studies}
\begin{figure}[h]
	\centering
	\includegraphics[width=\linewidth]{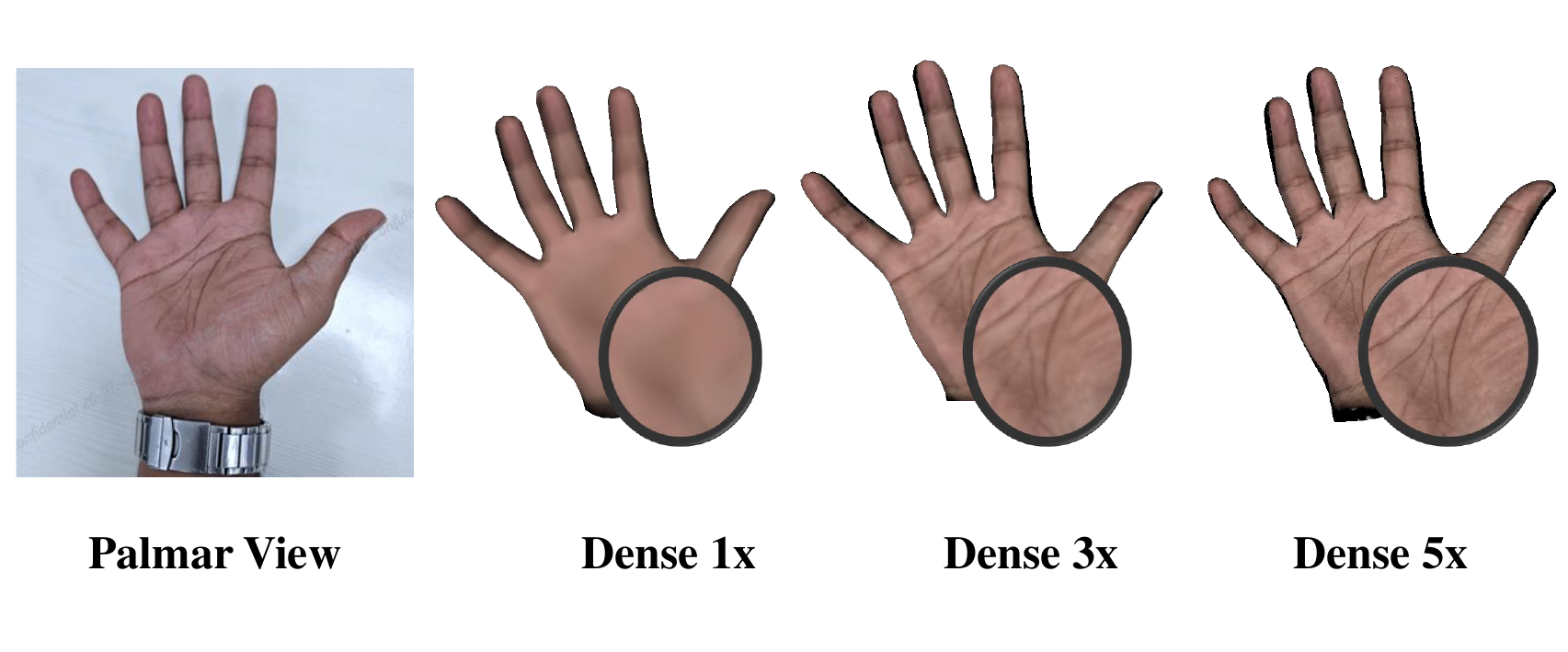}
	\caption{ Impact of mesh resolution. A dense mesh results in higher fidelity outputs (circle regions).}
	\label{fig: densification}
\end{figure}

To assess the role of geometric density in high-fidelity hand texture synthesis, we conduct an ablation study by subdividing the MANO \cite{manoref} upto 5$\times$, which produced significantly higher DINOv2 \cite{dinov2ref} cosine similarity and ViTScore \cite{vitref} values as can be seen in Table \ref{table: densification_ablation}. We attribute this to denser meshes projecting more pixels per surface region, thereby enabling extraction of high-frequency texture details such as subtle skin patterns as shown in Fig. \ref{fig: densification}.  However, densification increases computational cost, texture generation time doubles (from $\sim$ 38s to $\sim$ 76s, as shown in Table. \ref{table: densification_ablation} Col 4)  due to finer fragment processing. The trade-off is justified as distinctive markers must remain sharp. These results confirm that mesh densification amplifies perceptual quality and improves semantic similarity scoring, validating its integration in our workflow.

\begin{figure}[h]
	\centering
	\includegraphics[width=\linewidth]{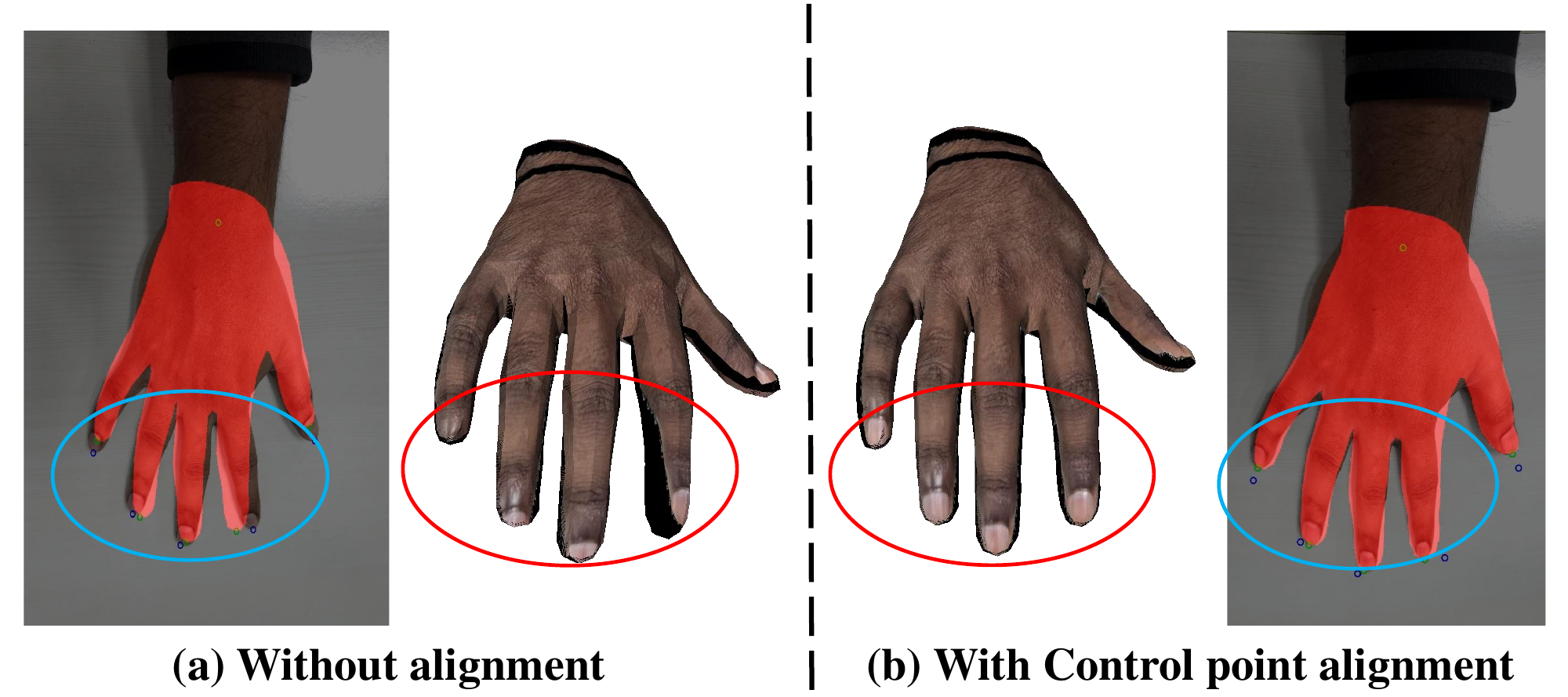}
	\caption{Control point alignment (b) results in a much better textured mesh, with less holes compared to the output without alignment (a).}
	\label{fig: ablation_1}
\end{figure}

We also evaluated the impact of Thin Plate Spline (TPS) \cite{tpsref} warping for mesh–image alignment during texture extraction. We compared 4 configurations: No alignment, control anchors with 6 points, 14 points and 18 points. Employing 14 control points yielded the most precise registration between the mesh and hand image, ensuring anatomical features like fingernail tips align correctly, though processing time roughly tripled relative to the unaligned baseline. This was needed as proper projection alignment is crucial for accurate texture capture; without it, misplacements can occur, for example, nails may not appear at fingertip positions, resulting in unnatural artifacts. Fig. \ref{fig: ablation_1} illustrates this effect, showing how TPS \cite{tpsref} alignment mitigates misregistration and permits extraction of high-fidelity, identity-consistent texture details. However as the number of control points become big, the warping space becomes highly nonlinear resulting in twisting artefacts in the warped image, hence the texture on the hand appears distorted, this reflects as low semantic scores as shown in the Table. \ref{table: control_points_ablation} Row 4.

\begin{table}[h]
\caption{Ablation Study: Texture Fidelity with increased densification}
\resizebox{0.95\linewidth}{!}{%
\begin{tabular}{c|c|c|c}
\hline
\multicolumn{1}{c|}{\textbf{Densification}} & \multicolumn{1}{c|}{\textbf{ViT \cite{vitref} Score}} & \multicolumn{1}{c|}{\textbf{DINOv2 \cite{dinov2ref} Score}} & \multicolumn{1}{c}{\textbf{Time (in sec)}} \\ \hline
Dense 1x & 0.739 & 0.595 & 76.26 \\ \hline
Dense 3x & 0.810 & 0.717 & 40.38 \\ \hline
Dense 5x & \textbf{0.891} & \textbf{0.896} & \textbf{38.98} \\ \hline
\end{tabular}%
}
\label{table: densification_ablation}
\end{table}

\begin{table}[h]
\caption{Ablation Study: Texture Fidelity with number of control points}
\resizebox{0.95\linewidth}{!}{%
\begin{tabular}{c|c|c|c}
\hline
\multicolumn{1}{c|}{\textbf{Control Points}} & \multicolumn{1}{c|}{\textbf{ViT \cite{vitref} Score}} & \multicolumn{1}{c|}{\textbf{DINOv2 \cite{dinov2ref} Score}} & \textbf{Time} \\ \hline
No Control Points & 0.705 & 0.773 & 28.838 \\ \hline
6 Control Points & 0.789 & 0.885 & 63.516 \\ \hline
14 Control Points & \textbf{0.891} & \textbf{0.896} & 76.259 \\ \hline
18 Control Points & 0.524 & 0.432 & 90.259 \\ \hline
\end{tabular}%
}
\label{table: control_points_ablation}
\end{table}
\vspace{-0.15cm}
\section{Conclusion}
In this work, we present an integrated framework for high fidelity hand texture synthesis that combines semantically guided mesh alignment, densified texture extraction, and controlled domain-specific inpainting to synthesize personalised photo-realistic multi-view renders. With this work we paved the path for developing an optimisation free, fast pipeline to generate textures on the fly by using just frontal and dorsal images with high fidelity visual textures. Though our texture extraction was done using high resolution mesh to extract fine details, our final rendering uses a low-resolution mesh ,with pre-computed normals and a single texture map with efficient runtime performance. Our method employs optimized design strategies to create fast textures while maintaining visual quality, ideal for AR/VR applications.

\bibliographystyle{ACM-Reference-Format}
\bibliography{PHAF}

\appendix

\end{document}